\documentclass[sigconf]{acmart}

\usepackage{booktabs} 
\usepackage{balance}
\usepackage{subcaption,multirow}
\usepackage{tikz}
\usetikzlibrary{arrows}
\usetikzlibrary{shapes}
\usepackage[edges]{forest}
\usetikzlibrary{shadows.blur}
\usetikzlibrary{shapes.multipart}
\usepackage{todonotes}

\AtBeginDocument{%
  \providecommand\BibTeX{{%
    \normalfont B\kern-0.5em{\scshape i\kern-0.25em b}\kern-0.8em\TeX}}}
\setcopyright{acmcopyright}
\copyrightyear{2020}
\acmYear{2020}
\acmConference[GECCO '20]{Genetic and Evolutionary Computation Conference}{July 8--12, 2020}{Cancún, Mexico}
\acmBooktitle{Genetic and Evolutionary Computation Conference (GECCO '20), July 8--12, 2020, Cancún, Mexico}
\acmPrice{15.00}
\acmDOI{10.1145/3377930.3389838}
\acmISBN{978-1-4503-7128-5/20/07}

\usepackage{graphicx,xcolor,url}
\title{Versatile Black-Box Optimization}

\author{Jialin Liu}
\affiliation{
\institution{Southern University of Science and Technology}\city{Shenzhen, China}}
\email{liujl@sustech.edu.cn}
\author{Antoine Moreau}
\affiliation{\institution{Université Clermont Auvergne, CNRS, SIGMA Clermont, Institut Pascal}\city{Clermont-Ferrand, France}}
\email{antoine.moreau@univ-bpclermont.fr}
\author{Mike Preuss} 
\affiliation{\institution{LIACS, Universiteit}\city{Leiden, The Netherlands}}
\email{m.preuss@liacs.leidenuniv.nl}
\author{Jeremy Rapin, Baptiste Roziere}
\affiliation{
\institution{Facebook AI Research
\& Paris-Dauphine University}\city{Paris, France}
}\email{jrapin@fb.com,broz@fb.com}
\author{Fabien Teytaud}
\affiliation{
\institution{Univ. Littoral Cote d'Opale}\city{Calais, France}
}\email{teytaud@univ-littoral.fr}
\author{Olivier Teytaud}
\affiliation{\institution{Facebook AI Research}\city{Paris, France}
}
\email{oteytaud@fb.com}

\renewcommand{\shortauthors}{J. Liu, A. Moreau, M. Preuss, B. Roziere, F. Teytaud, J. Rapin, O. Teytaud} 

 \newcommand{\myedit}[1]{\textcolor{black}{#1}}
 \newcommand{\revfour}[1]{\textcolor{black}{#1}}

 \newcommand{\otc}[1]{\textcolor{black}{#1}}
 \newcommand{\otcdeux}[1]{\textcolor{black}{#1}}
 \newcommand{\br}[1]{\textcolor{black}{#1}}
  \newcommand{\jlr}[1]{\textcolor{black}{#1}}
\newcommand{\mike}[1]{\textcolor{black}{#1}}
\newcommand{\antoine}[1]{\textcolor{black}{#1}}
 \def\NGO{Shiwa}
\date{November 2019}

\copyrightyear{2020} 
\acmYear{2020} 
\setcopyright{acmcopyright}
\acmConference[GECCO '20]{Genetic and Evolutionary Computation Conference}{July 8--12, 2020}{Cancún, Mexico}
\acmBooktitle{Genetic and Evolutionary Computation Conference (GECCO '20), July 8--12, 2020, Cancún, Mexico}
\acmPrice{15.00}
\acmDOI{10.1145/3377930.3389838}
\acmISBN{978-1-4503-7128-5/20/07}

\begin{document}
\begin{abstract}
Choosing automatically the right algorithm using problem descriptors is a classical \otc{component of} combinatorial optimization. It is also a good tool for making evolutionary algorithms fast, robust and versatile. We present \NGO{}, an algorithm good at both discrete and continuous, noisy and noise-free, sequential and parallel, black-box optimization.
Our algorithm is \myedit{experimentally compared to competitors} on YABBOB\myedit{, a BBOB comparable} testbed, and on some variants of it, and then validated on several real world testbeds.
\end{abstract}

\ccsdesc[500]{Theory of computation~Optimization with randomized search heuristics}

\keywords{Black-box optimization, portfolio algorithm, gradient-free algorithms, open source platform}

\maketitle
\section{Introduction: algorithm selection}\label{pd}
Selecting automatically the right algorithm is critical for success \otc{in combinatorial optimization: our goal is to investigate how this can be applied in derivative-free optimization}. Algorithm selection~\cite{rice1976,smithmiles2009,kotthoff2014,bischl2016_aslib,kerschke2018bbob,kerschke2018survey} can be made a priori or dynamically. In the dynamic case, the idea is to take into account preliminary numerical results, i.e. starting with several optimizers concurrently and, afterwards, focusing on the best only~\cite{mersmann2011,pitzer2012,malan2013,munoz2015_as,cauwet2016algorithm,kerschke2018survey}. Algorithm selection among a portfolio of methods routinely wins SAT competitions.
Passive algorithm selection~\cite{competencemap} is the special case in which the algorithm selection is made a priori, from high level characteristics which are typically known in advance, such as the following problem descriptors:
\myedit{dimension; computational budget \otc{(measured in terms of number of fitness evaluations)}; degree of parallelism in the optimization; nature of variables, whether they are discrete or not, ordered or not, whether the domain is metrizable or not\footnote{\otc{We consider that a domain is metrizable if it is equipped with a meaningful, non-binary, metric. In the present paper, any domain containing a discrete categorical variable is considered as non-metrizable. This is not the mathematical notion of metrizability.}}; presence of constraints; noise in the objective function; and multiobjective nature of the problem.}
\def\removedtosavespace{
\begin{itemize}
    \item Dimension.
    \item Computational budget (measured in terms of number of fitness evaluations).
    \item \revfour{Parallelism in the optimization: is it parallel, or does it have $p$ function evaluations running in parallel ?}
    \item Nature of variables, whether they are discrete or not, ordered or not, whether the domain is metrizable or not.\footnote{\otc{We consider that a domain is metrizable if it is equipped with a meaningful, non-binary, metric. In the present paper, any domain containing a discrete categorical variable is considered as non-metrizable. This is not the mathematical notion of metrizability.}}
    \item Presence of constraints.
    \item Noise in the objective function.
    \item Multiobjective nature of the problem.
\end{itemize}}
Besides the performance improvement on a \revfour{given   family of optimization problems}, algorithm selection also aims at designing a versatile algorithm, i.e. an algorithm which works for arbitrary domains and goals.
We use \mike{benchmark} 
experiments in \revfour{an extendable open source platform~\cite{nevergrad}, which contains a collection of benchmark problems and state of the-art gradient-free algorithms,} for designing a vast algorithm selection \otc{process} - and test it on real world objective functions.
\mike{Up to our knowledge, there is no direct competitor we could compare to because algorithm selection is usually only performed on a subset of the problem types we tackle. \otc{However, we do outperform the native algorithm selection methods of Nevergrad (CMandAS and CMandAS2~\cite{gecco2019}) on YABBOB (Fig. \ref{fig:oriyabbob}), though not by a wide margin as on testbeds far from their usual context (Fig. \ref{fig:axp}).}}
\section{Design of \NGO{}}
We propose \NGO{} \br{for algorithm selection}. \br{It uses a vast collection of algorithm selection \otc{rules} for optimization. Most of these \otc{rules} are passive, but some are active.} 
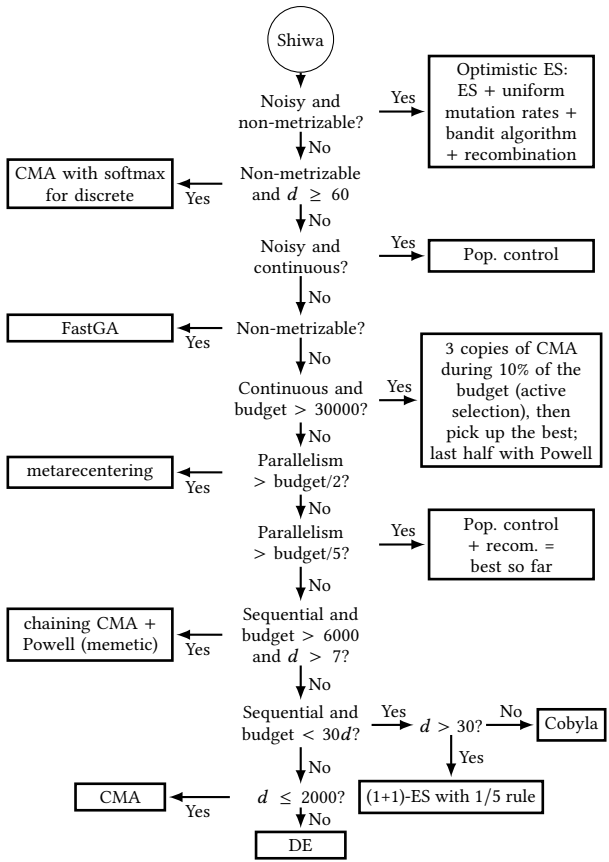
\begin{figure}[bhp]
\begin{center}\small
\begin{tikzpicture}[-,>=latex,shorten >=1pt,auto,node distance=2cm,font=\footnotesize, main node/.style={circle,draw},scale=.8,text centered]
\draw(0,-3.6) node(O) [circle,draw] {Shiwa};
\draw(0,-4.8) node(C1) [inner sep=1pt,text width=2cm] {Noisy and non-metrizable?};
\draw(0,-6) node(C2) [inner sep=1pt,text width=2cm] {Non-metrizable and $d\ge 60$};
\draw(0,-7.2) node(C4) [inner sep=1pt,text width=2cm] {Noisy and continuous?};
\draw(0,-8.4) node(C5) [inner sep=1pt,text width=2cm] {Non-metrizable?};
\draw(0,-9.6) node(C6) [inner sep=1pt,text width=2cm] {Continuous and budget $>30000$?};
\draw(0,-10.8) node(C7) [inner sep=1pt,text width=2cm] {Parallelism $>$ budget/2?};
\draw(0,-12) node(C8) [inner sep=1pt,text width=2cm] {Parallelism $>$ budget/5?};
\draw(0,-13.5) node(C10) [inner sep=1pt,text width=2cm] {Sequential and budget $>6000$ and $d>7$?};
\draw(0,-15) node(C9) [inner sep=1pt,text width=1.8cm] {Sequential and budget $<30d$?};
\draw(2.5,-15) node(C91) [inner sep=1pt] {$d>30$?};
\draw(0,-16.2) node(C11) [inner sep=1pt,text width=1.8cm] {$d\le 2000$?};

\path[->,draw,thick,black] (O) edge [out=270, in=90] (C1);

\draw(3.5,-4.8) node(A1) [rectangle, line width=1pt, draw,text width=2cm] {\revfour{Optimistic ES:} ES $+$ uniform mutation rates $+$ bandit algorithm $+$ recombination};
\draw(-3.5,-6) node(A2) [rectangle, line width=1pt, draw,text width=2cm] {CMA with softmax for discrete};
\draw(3.5,-7.2) node(A4) [rectangle, line width=1pt, draw,text width=2cm] {Pop. control};
\draw(-3.5,-8.4) node(A5) [rectangle, line width=1pt, draw,text width=2cm] {FastGA};
\draw(3.5,-9.6) node(A6) [rectangle, line width=1pt, draw,text width=2.2cm] {3 copies of CMA during $10\%$ of the budget (active selection), then pick up the best;\\ last half with Powell};
\draw(-3.5,-10.8) node(A7) [rectangle, line width=1pt, draw,text width=2cm] {metarecentering};
\draw(3.5,-12) node(A8) [rectangle, line width=1pt, draw,text width=2cm] {Pop. control $+$ recom. = best so far};
\draw(-3.5,-13.5) node(A10) [rectangle, line width=1pt, draw,text width=2cm] {chaining CMA $+$ Powell (memetic)};
\draw(4.5,-15) node(A91) [rectangle, line width=1pt, draw] {Cobyla};
\draw(2.5,-16.2) node(A92) [rectangle, line width=1pt, draw] {(1+1)-ES with $1/5$ rule};
\draw(-3,-16.2) node(A111) [rectangle, line width=1pt, draw,text width=1cm] {CMA};
\draw(0,-17) node(A112) [rectangle, line width=1pt, draw,text width=1cm] {DE};

\path[->,draw,thick,black] (C1) edge [out=0,in=180] node{Yes} (A1);
\path[->,draw,thick,black] (C1) edge [out=270,in=90] node{No} (C2);
\path[->,draw,thick,black] (C2) edge [out=180,in=0] node{Yes} (A2);
\path[->,draw,thick,black] (C2) edge [out=270,in=90] node{No} (C4);
\path[->,draw,thick,black] (C4) edge [out=0,in=180] node{Yes} (A4);
\path[->,draw,thick,black] (C4) edge [out=270,in=90] node{No} (C5);
\path[->,draw,thick,black] (C5) edge [out=180,in=0] node{Yes} (A5);
\path[->,draw,thick,black] (C5) edge [out=270,in=90] node{No} (C6);
\path[->,draw,thick,black] (C6) edge [out=0,in=180] node{Yes} (A6);
\path[->,draw,thick,black] (C6) edge [out=270,in=90] node{No} (C7);   
\path[->,draw,thick,black] (C7) edge [out=180,in=0] node{Yes} (A7);
\path[->,draw,thick,black] (C7) edge [out=270,in=90] node{No} (C8);
\path[->,draw,thick,black] (C8) edge [out=0,in=180] node{Yes} (A8);
\path[->,draw,thick,black] (C8) edge [out=270,in=90] node{No} (C10);   
\path[->,draw,thick,black] (C10) edge [out=180,in=0] node{Yes} (A10);
\path[->,draw,thick,black] (C10) edge [out=270,in=90] node{No} (C9);
\path[->,draw,thick,black] (C9) edge [out=0,in=180] node{Yes} (C91);
\path[->,draw,thick,black] (C9) edge [out=270,in=90] node{No} (C11);
\path[->,draw,thick,black] (C91) edge [out=0,in=180] node{No} (A91);
\path[->,draw,thick,black] (C91) edge [out=270,in=90] node{Yes} (A92);
\path[->,draw,thick,black] (C11) edge [out=180,in=0] node{Yes} (A111);
\path[->,draw,thick,black] (C11) edge [out=270,in=90] node{No} (A112);
\end{tikzpicture}
    \caption{\jlr{The \NGO{} algorithm \otc{using} the components listed in Section \ref{sec:components} \revfour{based on the hypotheses presented in Section \ref{prelim}}. The problem dimension is denoted by $d$.}}
    \label{medusa}
\end{center}
\end{figure}

\subsection{Components}\label{sec:components}
Our basic tools are \otc{methods from \cite{nevergrad}:}
\begin{itemize}
    \item Continuous domains: evolution strategies (ES)~\cite{Beyer:bookES} \myedit{including the Covariance Matrix Adaptation ES~\cite{hansen1996adapting} (abbreviated as ``CMA'' in this paper)}, estimation of distribution algorithms (EDA)~\cite{muhlenbein1996eda,bosman2000expanding,pelikan2002survey}, Bayesian optimization~\cite{ego}, particle swarm optimization (PSO)~\cite{pso}, differential evolution (DE)~\cite{storn1997differential,crde}, sequential quadratic programming~\cite{artelyssqp}, Cobyla~\cite{cobyla} and Powell~\cite{powell}. 
    \item Discrete optimization: the classical $(1+1)$-evolutionary algorithm, Fast-GA~\cite{fastga} and uniform mixing of mutation rates~\cite{danglehre}. We also include some recombination operators~\otc{\cite{holland}}.
    \item Noisy optimization:
bandits~\cite{bubeck}, algorithms with repeated sampling~\cite{decocknoise} and population control~\cite{beyerhellwignoise}.
\end{itemize}
\subsection{Combinations of algorithms}
Various tools exist for combining existing optimization algorithms. Terminology varies depending on authors. We adopt the following definitions:
	{\bf{ Chaining:}} running an algorithm, then another, and so on, initializing each algorithm using the results obtained by previous ones. In the present paper, we refer to chaining when an algorithm is run for a part of the computational budget, and another algorithm is run for the rest. For example, a memetic algorithm~\cite{memetic} running an evolution strategy at the beginning and Powell's method~\cite{powell} afterwards is a form of chaining.
	{\bf{Passive algorithm selection~\cite{competencemap}: }}
	the decision is made at the beginning of the optimization run, before any evaluation is performed.
	{\bf{ Active algorithm selection:}} 
	preliminary results of several optimization algorithms can be used to select one of them.
	{\bf{ Splitting:}} the variables are partitioned into $k$ groups of variables $G_1,\dots,G_k$. Then an optimizer $O_i$ optimizes variables in $G_i$; there are $k$ optimizers running concurrently. 
	All optimizers see the same fitness values, \br{and, at each iteration, the \otc{candidate} is obtained by \otc{concatenating} the \otc{candidates} proposed by each optimizer.}
Our optimization algorithm uses all these combinations,\ \otc{ except splitting - though there are problems for which splitting looks like a good solution and should be used}. Most of the improvement is due to passive algorithm selection for moderate computational budgets, whereas asymptotically active algorithm selection becomes critical (in particular for multimodal cases) and chaining (combining an evolution strategy for early stages and fast mathematical programming techniques at the end) is an important tool for large computational budget.
\subsection{Preliminaries}
\subsubsection{Ask and tell and recommend}
\otc{In the} present document, we use a ``ask and tell'' presentation of algorithms, convenient for presenting combinations of methods.
\def\numask{numask}
Given an optimizer $o$, $o.ask$ returns a candidate to be evaluated next. $o.tell(x,v)$ informs $o$ that the value at $x$ is $v$. $o.\numask$ is the number of times $o.ask$ has been used.
$o.archive$ is the list of visited points, with, for each of them, the list of evaluations (note that a same candidate can have been evaluated more than once, typically for noise management).
$o.recommend$ provides an approximation of the optimum; this is typically the final step of an optimization run.
\subsubsection{Softmax transformations}\label{softmax}
\cite{nevergrad} uses a softmax transformation for converting a discrete optimization problem into a noisy continuous optimization problem; for example, a variable with 3 possible values $a$, $b$ and $c$, becomes a triplet of continuous variables $v_a$, $v_b$ and $v_c$. The discrete variable has value $v_a$ with probability $\exp(v_a)/(\exp(v_a)+\exp(v_b)+\exp(v_c))$. \revfour{Preliminary experiments show} that this simple transformation \revfour{can perform} well for \revfour{discrete} variables with more than 2 values.
\subsubsection{Optimism, pessimism, progressive widening}
Following \cite{igel,gamecec2,coulom2007pw,wam}, we use combinations of bandits, progressive growing of the search space and evolutionary computation as already developed in Nevergrad. \revfour{Such combinations are labelled ``Optimistic'' in Fig. \ref{medusa}.} We refer to \cite{nevergrad} for full details.

\def\oldversion{
\cite{igel} proposed a combination of evolutionary computation and uncertainty bounds.
Similarly to~\cite{gamecec2}, we combine it with progressive widening~\cite{coulom2007pw, wam}, using the function $PW(n)$, true when 
$\lfloor n^{1/3}\rfloor >\lfloor (n-1)^{1/3}\rfloor$.
Given an optimizer $o$, our progressively widening optimistic counterpart $pw_o$ of $o$ is as follows:
\begin{eqnarray*}
pw_o.ask &=& UCB(archive)\mbox{ if not }PW(\numask)\\
pw_o.ask &=& o.ask(archivelcb)\mbox{ if }PW(\numask)\\
pw_o.tell(x,v)&=& o.tell(x, v) \\
pw_o.recommend&=& o.recommend(archivelcb)
\end{eqnarray*}
where \otc{$UCB(data)$ is a classical upper confidence bound algorithm picking up the point in data with the maximum upper confidence bound},
$archivelcb$ stands for a version of the archive with the fitness value of $x$ replaced by the lower confidence bound (LCB) of the fitness values at $x$ in $archive$. For several optimizers $o$, the $pw_o$ optimizer, \otc{exists} in Nevergrad and is termed optimistic noisy $o$.}
\subsection{\br{Algorithm design}}
\subsubsection{\revfour{Hypotheses}}\label{prelim}
\revfour{
The design of our algorithm \NGO{} relies on the following hypotheses:
}
	\revfour{(i) The standard $(1+1)$-ES in continuous domains is reliable for low computational budget / high dimension. 
	(ii) DE is a well known baseline algorithm. 
	(iii) CMA performs well in moderate dimension~\cite{hansen1996adapting}, in particular for rotated ill-conditioned problems.
	(iv) Population control~\cite{beyerhellwignoise} \revfour{is designed for} continuous parameter optimization in high-dimension with a noisy optimization function. (v) In other noisy optimization settings one can use an \otc{``optimistic''} combination of bandit algorithms and evolutionary computation~\cite{igel,vasilfoga} \otc{as detailed above.}
	(vi) Population control can be used for increasing the population size in case of stagnation, even without noise (see \otc{the ``NaiveTBPSA'' code} in~\cite{nevergrad}).}
	(vii) The MetaRecentering~\cite{metarecentering} algorithm \revfour{wins benchmarks in many one-shot optimization problems in \cite{nevergrad}}: it is a combination of Hammersley sampling~\cite{hammersley}, scrambling~\cite{atanassov}, and automatic rescaling~\cite{metarecentering}.
	(vii) Continuous optimization algorithms combined with softmax \revfour{can be used} for discrete optimization in the case of an alphabet of cardinal $>2$, \otcdeux{according to our results in discrete settings (unpresented but available online\footnote{\url{http://dl.fbaipublicfiles.com/nevergrad/allxps/list.html}}),
	\otcdeux{though the case} of a discrete alphabet of cardinal $2$ is \revfour{classically} handled with classical discrete evolutionary methods~\cite{danglehre,fastga}. }
	(viii) \revfour{Active portfolios are designed for large computational budgets} \myedit{(simple active portfolio methods, CMandAS and CMandAS2, are ranked as the 1st and 3rd in Figure \ref{fig:YABIGBBOB})}\otcdeux{, with our proposed methods as only competitors}.
	(ix) Some algorithms require a computational budget linear in the dimension for their initialization, hence poor performance occurs when the computational budget is low compared to the dimension.

\subsubsection{Algorithm: \NGO{}}\label{NGO}
\revfour{Based on the hypotheses presented above, we designed an algorithm structured as a decision tree (see Fig. \ref{medusa}). Constants in this decision tree were modified using preliminary experiments on the YABBOB testbed (defined below): \NGO{} as shown in Fig. \ref{medusa} is the final result. The algorithms mentioned in the boxes of Fig. \ref{medusa} were preexisting in Nevergrad.}
Our \NGO{} algorithm has been merged into \br{Nevergrad}~\cite{nevergrad} and is therefore publicly visible there\footnote{\url{https://github.com/facebookresearch/nevergrad/blob/master/nevergrad/optimization/optimizerlib.py}}.
\section{Experiments}
\begin{figure}
    \centering
\includegraphics[width=.43\textwidth]{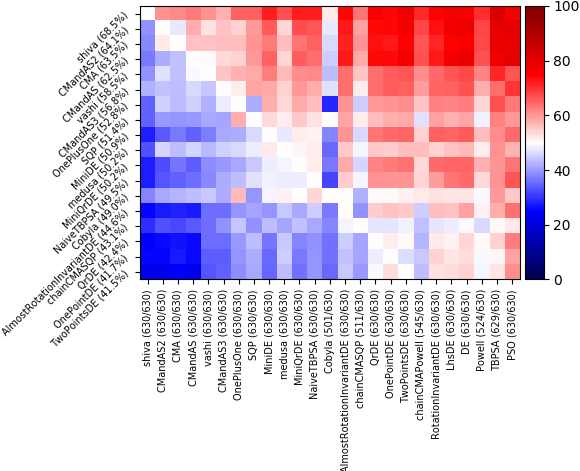}
\caption{YABBOB: \myedit{Sequential optimization with computational budget} $T\in \{ 50, 200, 800, 3200, 12800\}$. \otc{Reading guide: we see that \NGO{} performs better on average over all YABBOB experiments, than \otc{each of the other} algorithms (i.e. all colors in \NGO{}'s row are $\geq 50\%$). On average over all YABBOB experiments, \NGO{} wins with a probability of 68.5\% against other methods. ``winning'' means that the point recommended by \NGO{} had a better fitness value than the point recommended by another method (ties have truth value $\frac12$). CMandAS and CMandAS2 are simple active portfolio methods present in Nevergrad and fully described there.}}\label{fig:oriyabbob}
\end{figure}
\begin{figure*}
    \centering
\begin{subfigure}[t]{0.49\textwidth}\centering
\includegraphics[width=.8\textwidth]{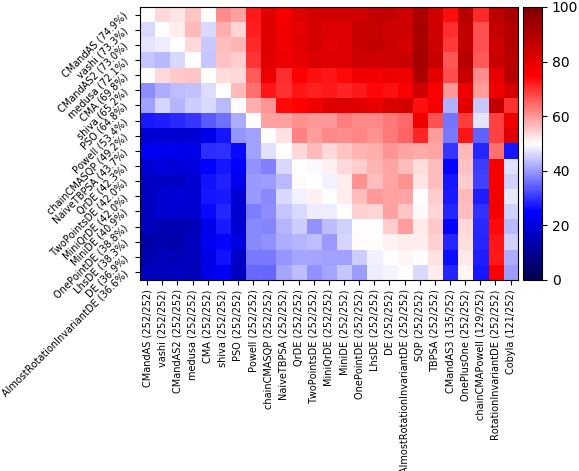}
	\caption{\label{fig:YABIGBBOB}YABIGBBOB: \otc{counterpart of YABBOB with} $T\in\{40000,80000\}$, $d\in\{2,10,50\}$. Vashi and Medusa are clones of \NGO{} sometimes present \otc{in} the \myedit{Nevergrad platform}.}
\end{subfigure}\hfill
\begin{subfigure}[t]{0.49\textwidth}\centering
\includegraphics[width=.8\textwidth]{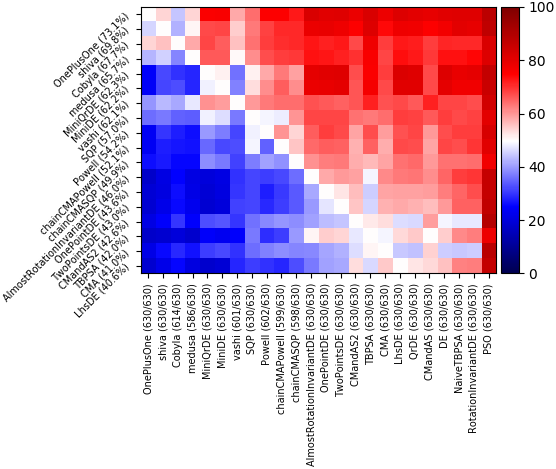}
\caption{YAHDBBOB: \otc{counterpart of YABBOB with} $d\in\{100, 1000, 3000\}$. \otc{CMA performs poorly in this case.}}
\end{subfigure}\\
\begin{subfigure}[t]{0.49\textwidth}\centering
\includegraphics[width=.8\textwidth]{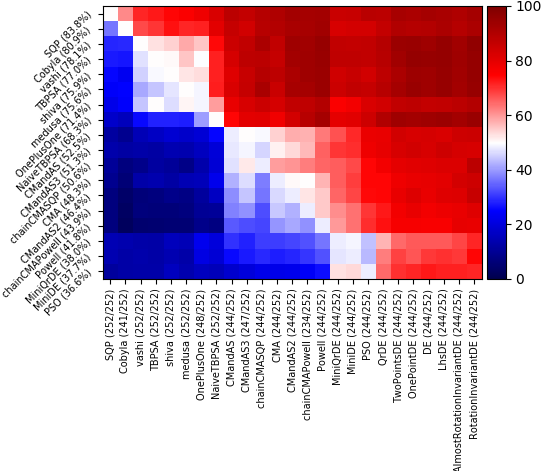}
\caption{YANOISYBBOB: \otc{counterpart of YABBOB with} \revfour{Gaussian noise} added, with variance not vanishing to zero at the optimum. \otc{\NGO{} is on par with methods good at noisy problems.}}
\end{subfigure}\hfill
\begin{subfigure}[t]{0.49\textwidth}\centering
\includegraphics[width=.8\textwidth]{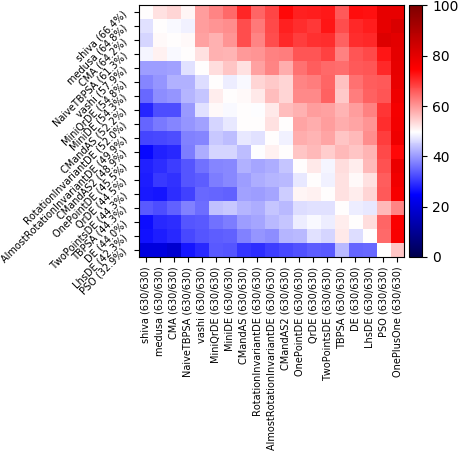}
	\caption{YAPARABBOB: \otc{counterpart of YABBOB with} 100 parallel workers. \otc{Medusa is a clone of \NGO{} sometimes present in the \myedit{Nevergrad platform}.}}
\end{subfigure}
\caption{\otc{Variants of YABBOB: YANOISYBBOB, YAHDBBOB, YAPARABBOB, YABIGBBOB, corresponding to the counterpart} one with noise, the high-dimensional one, the parallel one, the big computational budget, respectively. $T$ is the computational budget and $d$ refers to the dimension. \otc{\NGO{} is good in all categories, often performing the best, and always competing decently with the best in that category.}}
    \label{fig:axp}
\end{figure*}
\begin{figure*}[htbp]
\centering
\begin{subfigure}[t]{0.49\textwidth}\centering
\includegraphics[width=.8\textwidth]{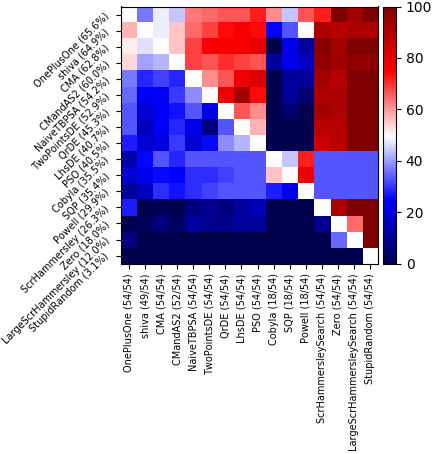}
	\caption{ARcoating: optical properties of layered structures. \otc{\NGO{} is outperformed by the simple $(1+1)$-ES; for the more general Photonics problem (covering a wider range of cases) this is not the case.} }
\end{subfigure}\hfill
\begin{subfigure}[t]{0.49\textwidth}\centering
\includegraphics[width=.8\textwidth]{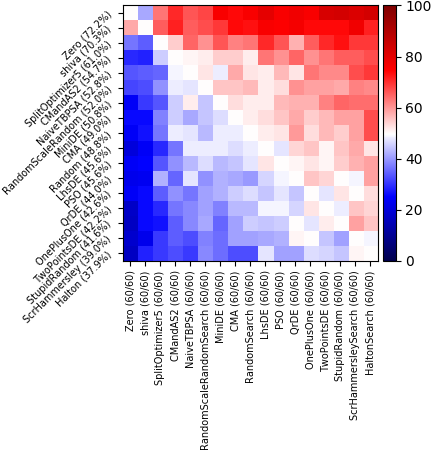}
\caption{FastGames (tuning strategies at real-world games; zero uses an already optimized policy, hence is not a real competitor.)}
\end{subfigure}
\def\letussaveupspace{
\begin{subfigure}[t]{0.49\textwidth}\centering
\includegraphics[width=.9\textwidth]{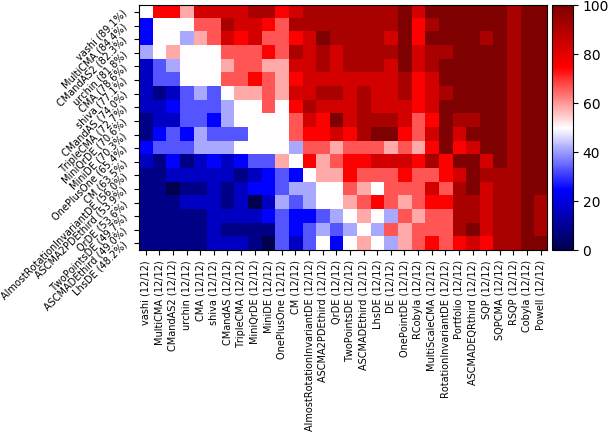}
	\caption{Constrained-Ill-\revfour{Conditioned}. \otc{\NGO{} does not use the presence of constraints as a feature descriptor. CMA is the only algorithm outperforming \NGO{} without being another form of combined algorithm. Vashi and Urchin are clones of \NGO{} available in the \myedit{Nevergrad platform} at that time. Using this feature descriptor is a further work.}}
\end{subfigure}\hfill
\begin{subfigure}[t]{0.49\textwidth}\centering
\includegraphics[width=1\textwidth]{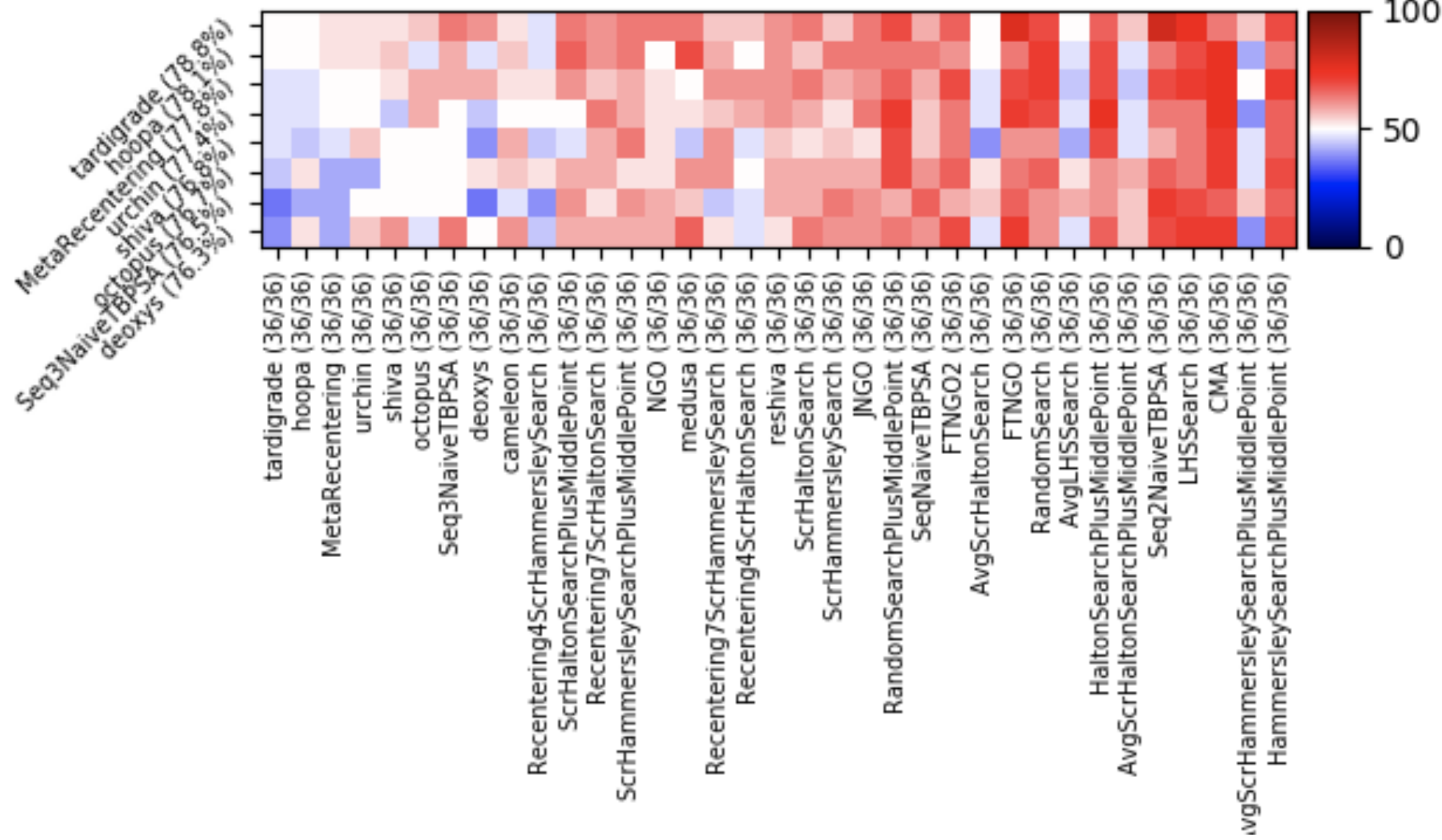}
	\caption{OneShot optimization: optimizing the sphere, the cigar and the Rastrigin function with 3 or 25 critical variables and 100\% or 16.67\% of critical variables with budget $T$ in 30, 100 or 3000. Octopus, NGO, Tardigrade, Hoopa, Urchin, Deoxys, are variants of \NGO{} that were in the \myedit{Nevergrad platform} at that time. \NGO{} boils down to MetaRecentering, hence the similar performance.}
\end{subfigure}\hfill
}
	\caption{Results on specific homogeneous families of problems (to be continued in Fig. \ref{fig:axp3}). \otc{\NGO{} performs well overall.}}
    \label{fig:axp2}
\end{figure*}
\begin{figure*}[htbp]
\centering
\begin{subfigure}[t]{0.49\textwidth}\centering
\includegraphics[width=.7\textwidth]{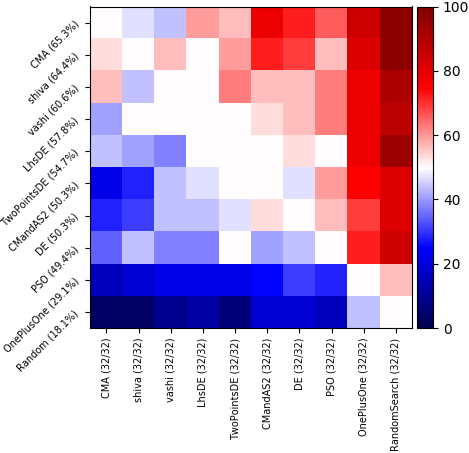}
\caption{Multiobjective optimization (set of problems with 2 or 3 objectives). \textcolor{black}{The negative hypervolume is used as fitness value for converting this multiobjective problem into a monoobjective one.} \textcolor{black}{See (5b) for discussion.}}
\end{subfigure}\hfill
\begin{subfigure}[t]{0.49\textwidth}\centering
\includegraphics[width=.7\textwidth]{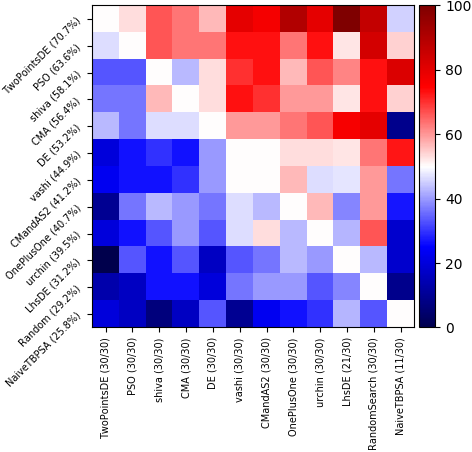}
\caption{Manyobjective optimization \otc{(set of problems with 6 objectives). \otcdeux{\NGO{} and CMA perform on par for these two multiobjective cases (5a) and (5b).}}}
\end{subfigure}
\begin{subfigure}[b]{0.49\textwidth}\centering
\includegraphics[width=.8\textwidth]{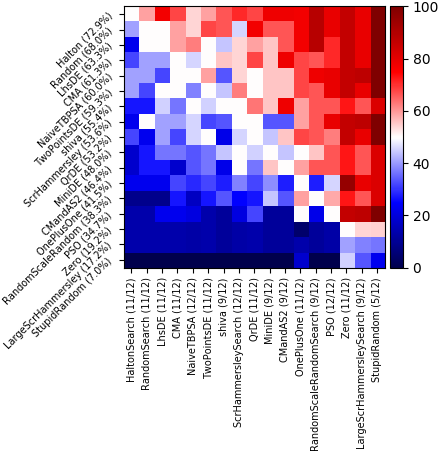}
\caption{MLDA\otc{: Machine Learning and Data Analysis}~\cite{mlda}.}
\end{subfigure}\hfill
\begin{subfigure}[b]{0.49\textwidth}\centering
\includegraphics[width=.9\textwidth]{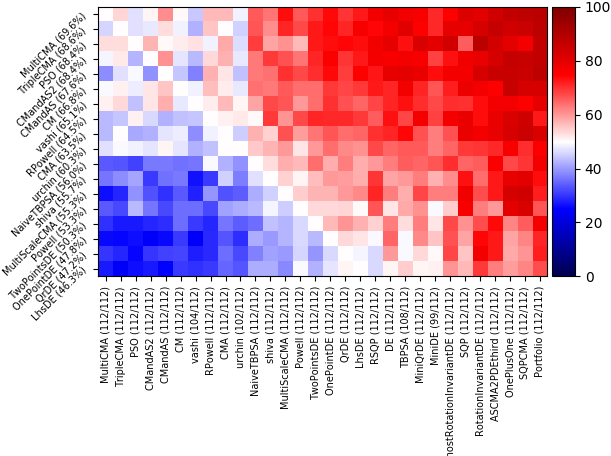}
\caption{Multimodal optimization\otc{: hm, Rastrigin, Griewank, Rosenbrock, Ackley, Lunacek, DeceptiveMultimodal; 3 or 25 critical variables; 0 or 5 useless variables per critical variable; $T\in \{ 3000, 10000, 30000, 100000 \}$}.}
\end{subfigure}
\caption{We here plot results on other specific homogeneous families of problems (Fig. \ref{fig:axp2} continued; to be continued in Fig. \ref{fig:axp4}).}
\label{fig:axp3}
\end{figure*}
\begin{figure*}[htbp]
\centering
\begin{subfigure}[t]{0.49\textwidth}\centering
\includegraphics[width=1\textwidth]{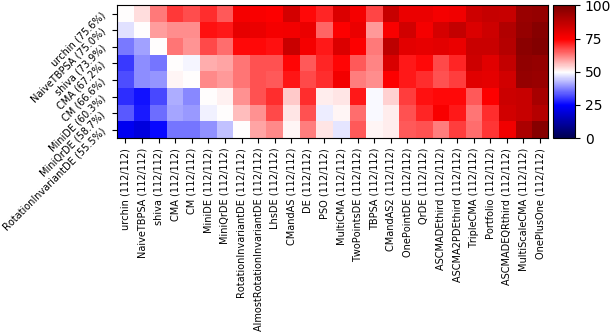}
\caption{Parallel Multimodal (1000 parallel workers).}
\end{subfigure}\hfill
\def\removethat{\begin{subfigure}[t]{0.35\textwidth}\centering
\includegraphics[width=.8\textwidth]{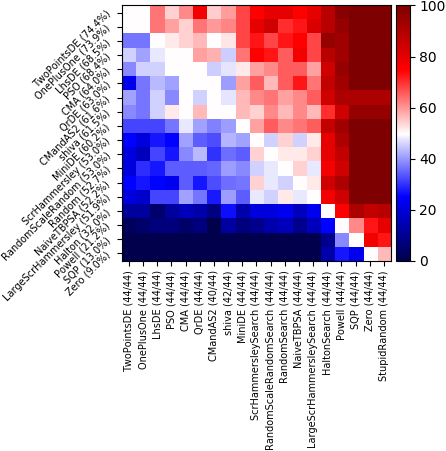}
\caption{SimpleTSP (traveling salesman problems).}
\end{subfigure}\\
}
\begin{subfigure}[t]{0.49\textwidth}\centering
\includegraphics[width=.7\textwidth]{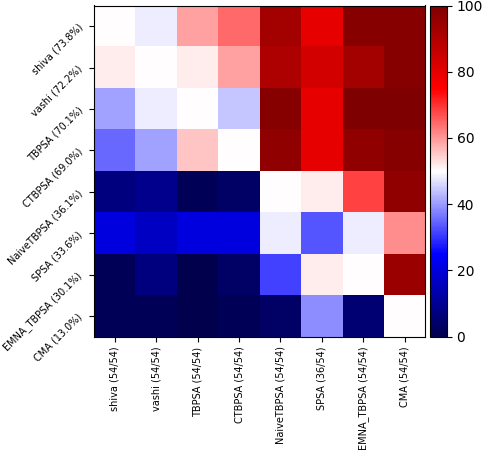}
\caption{SPSA test: noisy benchmark designed for testing \myedit{Simultaneous Perturbation Stochastic Approximation (SPSA)~\cite{spall}}.}
\end{subfigure}
\hfill
\def\removethat{
\begin{subfigure}[t]{0.49\textwidth}\centering
\includegraphics[width=.7\textwidth]{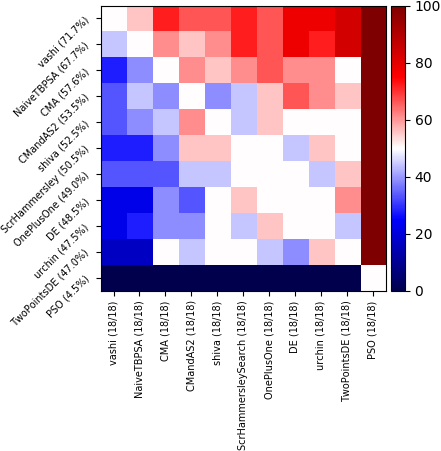}
\caption{Parallel artificial optimization\otc{: \#workers $=$ 20\% of the budget. Here \NGO{} is outperformed by CMA and by Vashi, a modified clone of \NGO{}, and by \myedit{NaiveTBPSA, a naive variant of Test-Based Population Size Adaptation~\cite{beyerhellwignoise}, discussed in the text}.}}
\end{subfigure}\\}
\caption{We here plot results on other specific homogeneous families of problems (Fig. \ref{fig:axp3} continued).}
    \label{fig:axp4}
\end{figure*}
\begin{figure}[htbp]
\includegraphics[width=.43\textwidth]{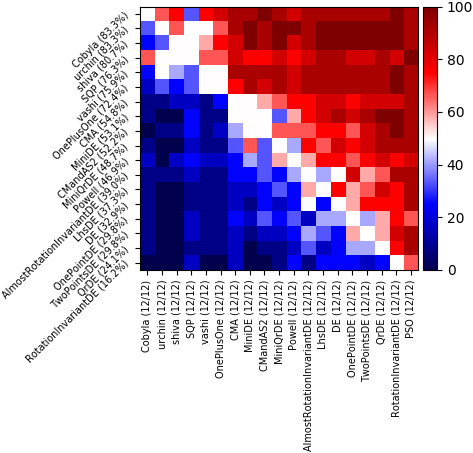}
\caption{Ill-conditioned optimization (``illcondi'' in~\cite{nevergrad}). Here this is a big gap between the top algorithms and others - evolutionary algorithms are not competitive on smooth ill-conditioned functions.\label{illc}}
\end{figure}

We compare \NGO{} to algorithms from Nevergrad on a large and diverse set of problems. The results of these experiments, which have all been produced using Nevergrad, 
\revfour{are summarized in Figs. \ref{fig:oriyabbob}, \ref{fig:axp}, \ref{fig:axp2}, \ref{fig:axp3}, \ref{fig:axp4} and \ref{illc}.} 
{\revfour{The  plots are made} as follows.} 

    A limited number of rows is presented (top for the best). Each row corresponds to an optimizer. Optimizers are ranked by their scores.
    \jlr{The score of an optimizer $o$ in a testbed ${\mathcal{T}}$ is the frequency, averaged over all problems $p\in {\mathcal{T}}$ and over all optimizers $o'$ included in the experiments, of $o$ outperforming $o'$ on $p\in {\mathcal{T}}$. The problems differ in the objective function, parametrization, computational budget $T$, dimension $d$, \otc{presence of random rotation or not, }and the degree of parallelization.} \otc{``$o$ outperforming $o'$ on a function $f$'' means that $f(o.recommend)<f(o'.recommend)$, with a truth value of $\frac12$ in case of tie.}
    Similarly to~\cite{HAFR2012RPBBOBES}, the comparison is made among implementations, not among abstract algorithms; the detailed implementation\otc{s are} freely available and open sourced in~\cite{nevergrad} or in modules imported there.
    The columns correspond to the same optimizers, but there are more optimizers included. Still, not all optimizers are shown. Between parentheses, we can see how many cases were run out of how many instances exist; e.g. 244/252 means that there are 252 instances, but only 244 were successfully run. A failure can be due to a bug, or, in most cases, to the timeout - some optimizers become computationally expensive in high-dimensional problems and fail to complete. 
\paragraph{\otcdeux{Detailed setup and reproducibility.}} The detailed setup \otc{of our benchmarks} is available online\footnote{ \url{https://github.com/facebookresearch/nevergrad/blob/master/nevergrad/benchmark/experiments.py}}. 
We did not change any setting. Our plots are those produced automatically \otc{and periodically }by Nevergrad.
\otc{In the artificial testbeds, Nevergrad considers cases with \otc{different} numbers of critical variables (i.e. variables which have an impact on the fitness functions) and of useless variables (possibly zero). There are also rotated cases and non-rotated cases. The detailed setup is at the URL above. We provide a high-level presentation of benchmarks in the present document.}
\paragraph{Statistical significance.} \revfour{Each experiment contains several settings.} The total number of repetitions varies, but there are always at least 200 \revfour{random repetitions} (cumulated over the different settings) for each result displayed in our plots, and at least 5 \revfour{repetitions} per setting. Importantly, Nevergrad periodically reruns all benchmarks with all algorithms: we can see independent reruns of all results presented in the present paper at \url{http://dl.fbaipublicfiles.com/nevergrad/allxps/list.html}, plus additional experiments (including a so-called YAWIDEBBOB which extends YABBOB) and results are essentially the same. 
\subsection{YABBOB: Yet Another Black-Box Optimization Benchmark}
YABBOB is a benchmark of black-box optimization problems. It roughly approximates BBOB~\cite{HAFR2012RPBBOBES}, without exactly sticking to it. \revfour{Consequently, CMA performs the best on it.}
YABBOB has the following advantages. It is 
part of a maintained and easy to use platform.
    Optimization algorithms are natively included in an open sourced platform.
 Everyone can rerun everything.
    There is a noisy optimization counterpart, without the issues known in BBOB~\cite{bbobissue1,bbobissue2,bbobissue3,bbobissue4}.
     There are several variants, parallel or not, noisy or not, high-dimensional or not, large computational budget or not.
     The same platform includes a large range of real world testbeds, switching form artificial to real world is just one parameter to change in the command line. 
\subsection{Training: artificial testbeds}
\revfour{We designed \NGO{} by handcrafting an algorithm as shown in Fig. \ref{medusa}, in which the constants have been tuned on YABBOB (Fig. \ref{fig:oriyabbob}) \otc{by trial-and-error}.
\revfour{Then,} variants of YABBOB, namely YANOISYBBOB, YAHDBBOB, YAPARABBOB (respectively noisy, high-dimensional, and parallel counterparts of YABBOB), \otc{presented in Fig. \ref{fig:axp},} and several other benchmarks (see details in Section \ref{oth}) are used for testing the generality of \NGO{}.} 
Tables \ref{thebb} and \ref{theotherbb} present our benchmarks.
\begin{table*}[htbp]
    \centering
        \caption{\revfour{YABBOB testbed and variants. The objective functions are Hm, Rastrigin, Griewank, Rosenbrock, Ackley, Lunacek, DeceptiveMultimodal, BucheRastrigin, Multipeak,
    Sphere, DoubleLinearSlope, StepDoubleLinearSlope,
    Cigar, AltCigar, Ellipsoid, AltEllipsoid, StepEllipsoid, 
    Discus, BentCigar, DeceptiveIllcond, DeceptiveMultimodal, DeceptivePath. Full details in \cite{nevergrad}. Reading guide: the YABBOB benchmark contains all these functions, tested in dimension 2, 10 and 50; with budget ranging from 50 to 12800; both in the rotated and not rotated case; without noise, and always with random translations $+\mathcal{N}(0,Id)$.}}
    \label{thebb}
   \setlength{\tabcolsep}{6pt}
    \begin{tabular}{ccccccc}
    \hline
    Name & Dimensions & Budget & Parallelism & Translated & Rotated & Noisy \\
    \hline
    YABBOB & 2, 10, 50 & 50, 200, 800, 3200, 12800 & 1 (sequential) & $+{\mathcal{N}}(0,Id)$ & $\{yes,no\}$ & No \\
    YABIGBBOB & 2, 10, 50 & 40000, 80000 & 1 (sequential) & $+\mathcal{N}(0,Id)$ & $\{yes,no\}$ & No \\
    YAHDBBOB & 100, 1000, 3000 & 50, 200, 800, 3200, 12800 & 1 (sequential) & $+{\mathcal{N}}(0,Id)$ & $\{yes,no\}$ & No \\
    YANOISYBBOB & 2, 10, 50 & 50, 200, 800, 3200, 12800 & 1 (sequential) & $+{\mathcal{N}}(0,Id)$ & $\{yes,no\}$ & Yes \\
    \hline
    \end{tabular}

\end{table*}
\begin{table*}[htbp]
\centering
\caption{\revfour{Other benchmarks used in this paper, besides YABBOB and its variants. Full details in \cite{nevergrad}.}}
\label{theotherbb}
\setlength{\tabcolsep}{6pt}
\revfour{
\begin{tabular}{ccl}
\hline
Category & Name & \multicolumn{1}{c}{Description} \\
\hline
\multirow{5}{*}{Real world}&
ARCoating & Anti-reflective coating optimization. \\
& Photonics & Optical properties: Bragg, Morpho and Chirped. \\
&FastGames & Tuning of agents playing the game of war, Batawaf, GuessWho, BigGuessWho Flip.\\
&MLDA & Machine learning and data analysis testbed~\cite{mlda}\\
&Realworld & Includes many of the above and others, e.g. traveling salesman problems.\\
\hline
\multirow{8}{*}{Artificial}&
 \multirow{2}{*}{Multiobjective} & 2 or 3 objective functions among Sphere, Ellipsoid, Hm and Cigar in dimension 6 or 7,\\
  &  & sequential or 100 workers, budget from 100 to 5900.\\
 & Manyobjective & Similar with 6 objective functions.\\
& \multirow{2}{*}{Multimodal} & Hm, Rastrigin, Griewank, Rosenbrock, Ackley, Lunacek, DeceptiveMultimodal \\
 & & in dimension 3 to 25, sequential, budget from 3000 to 100000\\
& Paramultimodal & Similar with 1000 workers.\\
& Illcondi & Cigar, Ellipsoid, dimension 50, budget 100 to 10000, both rotated and not rotated.\\
& SPSA & Sphere, Cigar, various translations, strong noise. \\
\hline
\end{tabular}
}
\end{table*}
In addition, we validate our results on a wide real world testbed in Section \ref{rwsec}.
\subsection{Illustrating: homogeneous families of benchmarks.}\label{oth}
We here consider tests which are not directly from YABBOB variants. Some functions are common to some of YABBOB; this section, as opposed to the next one, is not intended to be a completely independent test: these tests are either related to YABBOB or to real world tests from the next section.
This section illustrates the behavior of \NGO{} on various families of functions\otc{, for analysis purpose}\mike{s}. Results are presented in Fig.  \ref{fig:axp2}, \ref{fig:axp3} and \ref{fig:axp4} \otc{and \ref{illc}}. $T$ and $d$ in the captions refer to the computational budget and dimension, respectively. We keep the name of these experiments as in~\cite{nevergrad}. \otcdeux{The context of Fig. \ref{illc} (``illcondi'' problem in Nevergrad) is as follows: smooth ill-conditioned problems, namely Cigar and Ellipsoid, both rotated and unrotated, in dimension 50 with moderate computational budget (from 100 to 10000). Unsurprisingly, for these problems for which ruggedness is null and the challenge is to tackle ill-conditioning within a moderate computational budget, SQP and Cobyla outperform all methods based on random exploration: \NGO{} (and its variant Urchin) can compete, without knowing anything about the objective functions, because, based on the low computational budget it switches to a chaining of CMA and Powell or to Cobyla (depending on the budget, see Fig. \ref{medusa}), and not to classical evolutionary methods alone.}
\def\bobof{
\begin{verbatim}
@registry.register
def illcondi(seed: tp.Optional[int] = None) -> tp.Iterator[Experiment]:
    """Testing optimizers on ill cond problems.
    """
    seedg = create_seed_generator(seed)
    optims = ["NGO", "Shiva", "DiagonalCMA", "CMA", "PSO", "DE", "MiniDE", "QrDE", "MiniQrDE", "LhsDE", "OnePlusOne", "SQP", "Cobyla",
              "Powell", "TwoPointsDE", "OnePointDE", "AlmostRotationInvariantDE", "RotationInvariantDE"]
    functions = [
        ArtificialFunction(name, block_dimension=50, rotation=rotation) for name in ["cigar", "ellipsoid"] for rotation in [True, False]
    ]
    for optim in optims:
        for function in functions:
            for budget in [100, 1000, 10000]:
                yield Experiment(function, optim, budget=budget, num_workers=1, seed=next(seedg))
                \end{verbatim}}
\subsection{Testing: real world testbeds}\label{rwsec}
\begin{figure}[htbp]
\centering
\includegraphics[width=.45\textwidth]{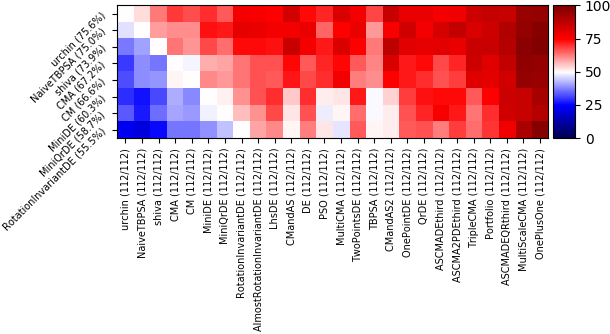}
\caption{Test on real-world problems, in the ``realworld'' benchmark from Nevergrad. This single figure aggregates many real world problems from~\cite{nevergrad} as documented there \otc{- here \NGO{} outperforms most competitors because it performs decently on all subfamilies of problems. This testbed contains 21 different functions, tested over 3 different degrees of parallelism, and many distinct budgets}. Urchin is a variant of \NGO{}.}\label{rwxp}
\end{figure}
\begin{figure}[htbp]
\centering
\includegraphics[width=.49\textwidth]{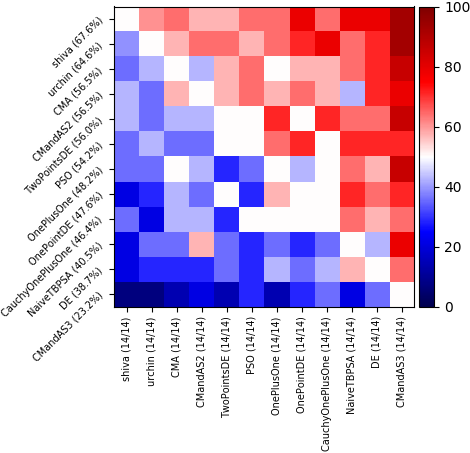}
\caption{\otc{Test on photonics problems, namely Bragg, Morpho and Chirped. This experiment is one of our contributions: it combines sequential and 50-parallel experiments, with dimension ranging from 60 to 80 and computational budget from 100 to 10000. Urchin is a slightly modified clone of \NGO{}.}}\label{photonics}
\end{figure}

We test our method on the ``realworld'' collection of problems in~\cite{nevergrad}.
This contains Traveling Salesman Problems (TSPs), Power Systems Management, 
\antoine{Photonics, ARCoating (design of anti-reflective coatings in optics~\cite{centeno2019ultra}),} Machine Learning and Data Applications (MLDA~\cite{mlda}), Games (policy optimization, incl. non-metrizable domains). Results are presented in Fig. \ref{rwxp}, and problems in this figure are all independent of YABBOB variants which were used for fine-tuning \NGO{}.
\antoine{Photonics and ARCoating problems are one of our contributions. The Photonics problems are truly evolutionary in the sense that the best solutions correspond to optical structures which occur in \otc{nature}~\cite{barry2018evolutionary} and have thus been produced by evolution. These testbeds are characterized by a very large number of local minima because these structures present potentially a lot of resonances. Finally, these problems are particularly modular, the different parts of the structures playing different roles without being truly independent.}

\section{Conclusions}
We designed \NGO{} (Section \ref{NGO}), a versatile optimization algorithm tackling optimization problems on arbitrary domains (continuous or discrete or mixed), noisy or not, parallel or oneshot or sequential, and outperforming many methods thanks to \br{a combination of a wide range of optimization algorithms.} 
The \otcdeux{remarks} in Section \ref{prelim}\otc{, namely comparisons between optimizers over various domains,} have scientific value by themselves, and are not visible in many existing testbeds. \otc{Experimental results include artificial ones: these results on YABBOB could be biased given that results on YABBOB were used for designing \NGO{}.} But the results are also  good on variants of YABBOB which were not used for designing the algorithms (Fig. \ref{fig:axp}), and on a wide real-world \otc{public testbed, namely ``realworld'' from Nevergrad, and on our new Photonics implementation}. \otc{Results were particularly good for wide families of problems, for which no homogeneous method alone can perform well.} 
\NGO{} can also take into account the presence of noise (without knowledge of the noise intensity, rarely available in real life), type of variables, computational budget, dimension, parallelism, for extracting the best of each method and outperforming existing methods. As an example of versatility, \NGO{} outperforms CMA on YABBOB, is excellent on real world benchmarks, \otc{on the new photonics problems,} \br{and} on ill-conditioned quadratic problems, and outperforms most one-shot optimization methods. \otc{In Fig. \ref{rwxp}, covering a wide range of problems which have never been used in the design of \NGO{}, \NGO{} performed well.} Reproducibility matters: all our code is integrated in~\cite{nevergrad} \br{and publicly available}.
\otc{
{\bf{Limitations.}}
We met classes of functions for which we failed to become better on the first, without degrading on the second. Typically, we are happy with the results in parallel, one-shot, high-dimensional or ill-conditioned cases: we get good results for any of these cases, without degrading more classical settings. The extension to noisy cases or discrete cases was also straightforward and without drawback. But for hard multimodal settings vs ill-conditioned unimodal functions, we did
our best to choose a realistic compromise but could not have something optimal for both. The no-free lunch~\cite{wolpert1997} actually tells us that we can not be universally optimal. Additional research in the active selection part might help.
}

{{\bf{Further work.}}}
1. We consider taking into account, as a problem feature, the computational cost of the objective function, the presence of constraints, and\otcdeux{, in particular for being more versatile regarding multimodal vs monomodal problems and rugged vs smooth problems,} using more active (i.e. online) portfolios\otcdeux{, or fitness analysis}. 2. A challenge is to recursively analyze the domain for combining different algorithms on different groups of variables: this is in progress.
3. Splitting algorithms are present in~\cite{nevergrad}. We feel that they are quite good in high dimension and could be part of \NGO{}. 4. \otc{Particle Swarm Optimization is excellent in some difficult cases: we try to leverage it by integrating it into \NGO{}. 5. \otcdeux{We did not present our experiments in discrete settings: they can however be retrieved in the section entitled ``wide-discrete'' of \url{http://dl.fbaipublicfiles.com/nevergrad/allxps/list.html}. We see there the good performance of tools using softmax, including \NGO{}. We did not include this in the present paper because, in the context of limited space, there was not enough diversity in the experiments. A thorough analysis of this method, rather new and beyond the scope of the present work dedicated to algorithm selection, is left as further work.
}}
6. \otcdeux{\NGO{} was mainly designed by scientific knowledge, and a bit by trial-and-error on YABBOB. We guess much better results could be obtained by a more systematic analysis, e.g. numerical optimization on a wider counterpart of YABBOB. 7. Integrating e.g. \cite{LGB17} for high-dimensional cases is also a possibility.}
\begin{acks}
Author J. Liu was supported by the National Key R\&D Program of China (Grant No. 2017YFC0804003), the National Natural Science Foundation of China (Grant No. 61906083), the Guangdong Provincial Key Laboratory (Grant No. 2020B121201001), the Program for Guangdong Introducing Innovative and Enterpreneurial Teams (Grant No. 2017ZT07X386), the Science and Technology Innovation Committee Foundation of Shenzhen (Grant No. JCYJ20190809121403553), the Shenzhen Science and Technology Program (Grant No. KQTD2016112514355531) and the Program for University Key Laboratory of Guangdong Province (Grant No. 2017KSYS008).
\end{acks}
\bibliographystyle{ACM-Reference-Format}
\balance
\bibliography{starNGO}
\end{document}